\documentclass{article}

\usepackage[preprint]{neurips_2019}

\usepackage[utf8]{inputenc} 
\usepackage[T1]{fontenc}    
\usepackage{hyperref}       
\usepackage{url}            
\usepackage{booktabs}       
\usepackage{amsfonts}       
\usepackage{nicefrac}       
\usepackage{microtype}      
\usepackage{amsmath}
\usepackage{subcaption}

\usepackage{pgfplots}
\pgfplotsset{compat=1.14}
\usepackage{tikz}
\usepackage{graphicx}

\title{Hierarchical Deep Q-Network from Imperfect Demonstrations in Minecraft}

\author{%
Alexey Skrynnik\\
Artificial Intelligence Research Institute FRC CSC RAS\\
\texttt{skrynnik@isa.ru} \\
\And
Aleksey Staroverov\\
Moscow Institute of Physics and Technology\\
\texttt{alstar8@yandex.ru} \\
\And
Ermek Aitygulov\\
Moscow Institute of Physics and Technology\\
\texttt{aytygulov@phystech.edu} \\
\And
Kirill Aksenov\\
Higher School of Economics\\
\texttt{kaaksyonov@edu.hse.ru} \\
\And
Vasilii Davydov\\
Moscow Aviation Institute\\
\texttt{dexfrost89@gmail.com} \\
\And
Aleksandr I. Panov\\
Artificial Intelligence Research Institute FRC CSC RAS\\
Moscow Institute of Physics and Technology\\
\texttt{panov.ai@mipt.ru} \\
}

\begin{document}

    \maketitle

    \begin{abstract}
        We present Hierarchical Deep Q-Network (HDQfD) that won first place in the MineRL competition.
        The HDQfD works on imperfect demonstrations and utilizes the hierarchical structure of expert trajectories.
        We introduce the procedure of extracting an effective sequence of meta-actions and subgoals from the demonstration data.
        We present a structured task-dependent replay buffer and an adaptive prioritizing technique that allow the HDQfD agent to gradually erase poor-quality expert data from the buffer.
        In this paper, we present the details of the HDQfD algorithm and give the experimental results in the Minecraft domain.
    \end{abstract}

    \section{Introduction}\label{sec:introduction}
    Deep reinforcement learning (RL) has achieved compelling success on many complex sequential decision-making problems, especially in simple domains.
    In such examples as AlphaStar~\cite{vinyals2019alphastar}, AlphaZero~\cite{silver2017mastering},
    OpenAI Five human or superhuman performance was attained.
    However, RL algorithms usually require a huge amount of environment-samples required for training to reach good performance~\cite{kakade2003sample}.
    Learning from demonstration is a well-known alternative, but until now, this approach has not achieved any considerable success in complex non-single-task environments.
    This was largely due to the fact that obtaining high-quality expert demonstrations in a sufficient quantity in sample-limited, real-world domains is a separate non-trivial problem.

    Minecraft as a compelling domain for the development of reinforcement and imitation learning based methods was recently introduced by~\cite{guss2019minerl}.
    Minecraft presents unique challenges because it is a 3D, first-person, open-world game where the agent should gather resources and create structures and items to achieve a goal.
    Due to its popularity as a video game it turned out to be possible to collect a large number of expert trajectories in which individual subtasks are solved.
    This allowed the appealing MineRL competition to run.
    The organizers have released the largest-ever dataset of human demonstrations on a Minecraft domain.
    The primary goal of the competition is to foster the development of algorithms that can efficiently leverage human priors to drastically reduce the number of samples needed to solve complex, hierarchical, and sparse environments.

    The main difficulty in solving the MineRL problem was the imperfection of demonstrations and the presence of hierarchical relationships of subtasks.
    In this paper we present hierarchical Deep Q-Network from Demonstrations (HDQfD) that allowed us to win first place in the MineRL competition proposed by~\cite{guss2019minerl2}.
    The HDQfD works on imperfect demonstrations and utilize a hierarchical structure of expert trajectories extracting effective sequence of meta-actions and subgoals.
    Each subtask is solved by its own simple strategy, which extends the DQfD approach proposed by~\cite{gao2018reinforcement} and relies on a structured buffer and gradually decrease the ratio of poor-quality expert data.
    In this paper, we present the details of our algorithm and provide the results that allow the HDQfD agent to play Minecraft at the human level.

    \section{Background}\label{sec:background}

    One way to explore the domain with the use of expert data is to do behavioral cloning (BC).
    Pure supervised learning BC methods suffer from a distribution shift, because the agent greedily imitates the demonstrated actions, it can drift away from the demonstrated states due to error accumulation.
    Another way to use expert data in search for the exploration policy is to use conventional RL methods like PPO, DDDQN, etc.
    and guide exploration through enforcing occupancy measure matching between the learned policy and current demonstrations.

    The main approach is to use demonstration trajectories sampled from an expert policy to guide the learning procedure by either putting the demonstrations into a replay buffer or using them to pre-train the policy in a supervised manner.

    The organizers of the MineRL competition provided a few baselines.
    Standard DQfD presented by~\cite{hester2018deep} get the max score of 64 after 1000 episodes, the PPO gets the max of 55 after 800 episodes, the rainbow also gets the max of 55 after 800 episodes of training.

    Our best solution exploits the method of injecting expert data into the agent replay buffer.
    The DQfD, which our method is based on, is an advanced approach to reinforcement learning with additional expert demonstrations.
    The main idea of the DQfD is to use an algorithm called Deep Q-Network (DQN) and combine loss function $J(Q)$, with the main component $J_E(Q)$:
    \begin{equation}
        J(Q) = J_{DQ}(Q) + {\lambda_1}J_n(Q) + {\lambda_2}J_E(Q) + {\lambda_3}J_{L2}(Q).
        \label{eq:losses}
    \end{equation}
    The loss function $J_{DQ}(Q)$ is a standard TD-error:
    \begin{equation}
        J_{DQ}(Q) = \left( R(s,a) + \gamma Q(s_{t+1}, a^{\max}_{t+1}; \theta') - Q(s, a; \theta) \right)^2.\label{eq:td}
    \end{equation}
    The loss function $J_n(Q)$ is the so-called N-step return, which allows the agent to extend the utility of trajectories to several steps, which lead to a better strategy:
    \begin{equation}
        J_n(Q) = r_t + {\gamma}r_{t+1} + \dots + \gamma^{n-1}r_{t+n-1} + max_{a}\gamma^{n}Q(s_{t+n}, a).\label{eq:n-step}
    \end{equation}
    The main part $J_E(Q)$ is a margin loss function.
    It is responsible for copying expert behavior and gives a penalty to the agent for performing actions other than those of the experts:
    \begin{equation}
        J_E(Q) = \max_{a \in A}[Q(s, a) + l(a_E, a)] - Q(s, a_E).\label{eq:margin}
    \end{equation}
    Finally, $J_{L2}(Q)$ is L2 regularization added to prevent overfitting.

    The problem of suboptimal demonstrations is considered by~\cite{nair2018overcoming}, who attempted to develop an approach closely related to ours.
    The authors use the BC loss as a margin loss only for expert data.
    The key idea of the suggested approach is Q-Filter, which applies the BC loss only to the states where the critic determines that the action of the demonstrator is better than that of the actor.
    The use of expert demonstrations depends on the quality of the critic, which can lead to difficulties in the case of sparse rewards.
    For such tasks, the authors of~\cite{nair2018overcoming} propose resets to the demonstration states technique for Hindsight Experience Replay~\cite{andrychowicz2017hindsight}.
    In contrast to the Q-filter, the HDQfD is suitable for solving a wide range of problems related to the quality of demonstrations.
    In this paper, our approach shows an ability to learn from expert data to which a change in the action space is applied.

    \section{Hierarchical Deep Q-Network from Demonstrations}\label{sec:hierarchical-deep-q-network-from-demonstrations}

    \subsection*{Action and state space}
    To make the demonstration data convenient for the RL agent we used action discretization and some techniques for state space preparation, in particular a frameskip and a framestack.
    The frameskip technique repeats the action selected by the agent over several steps.
    The framestack technique replaces the current state with the concatenation of the current and several previous states of the environment.
    In the MineRL simulator, the agent could choose between 10 actions (see Table~\ref{tab:discretization}).
    The expert action is mapped to the agent's action in the order shown in Table~\ref{tab:discretization}.
    For example, ``turn the camera right 10 degrees, turn the camera up 5 degrees, run forward'' will be mapped with the first action, i.e., turn the camera right 5 degrees and attack.
    All ``move''-actions (back, forward, left, right) were allowed because the experts used mostly them to direct the camera to the tree block.

    \begin{table}[ht]
        \caption{Discretization of actions used for all subtasks with frameskip 4.
        The expert action is mapped to the agent’s action in the order shown in this table.
        The rotation angle is determined using the sum of 4 frames.
        For other actions, the most frequent was selected.}
        \label{tab:discretization}
        \centering
        \begin{tabular}{lllllllllll}
            \toprule

            actions  & $a_0$ & $a_1$ & $a_2$ & $a_3$ & $a_4$ & $a_5$ & $a_6$ & $a_7$ & $a_8$ & $a_9$ \\
            \midrule
            pitch +5 & +     &       &       &       &       &       &       &       &       &       \\
            pitch -5 &       & +     &       &       &       &       &       &       &       &       \\
            yaw +5   &       &       & +     &       &       &       &       &       &       &       \\
            yaw -5   &       &       &       & +     &       &       &       &       &       &       \\
            forward  &       &       &       &       & +     & +     &       &       &       &       \\
            left     &       &       &       &       &       &       & +     &       &       &       \\
            right    &       &       &       &       &       &       &       & +     &       &       \\
            back     &       &       &       &       &       &       &       &       & +     &       \\
            jump     &       &       &       &       &       & +     &       &       &       & +     \\
            attack   & +     & +     & +     & +     & +     & +     & +     & +     & +     & +     \\
            \bottomrule
        \end{tabular}
    \end{table}

    \subsection*{Adaptive ration of expert data}
\begin{figure}[ht]
        \centering
        \includegraphics[width=0.9\linewidth]{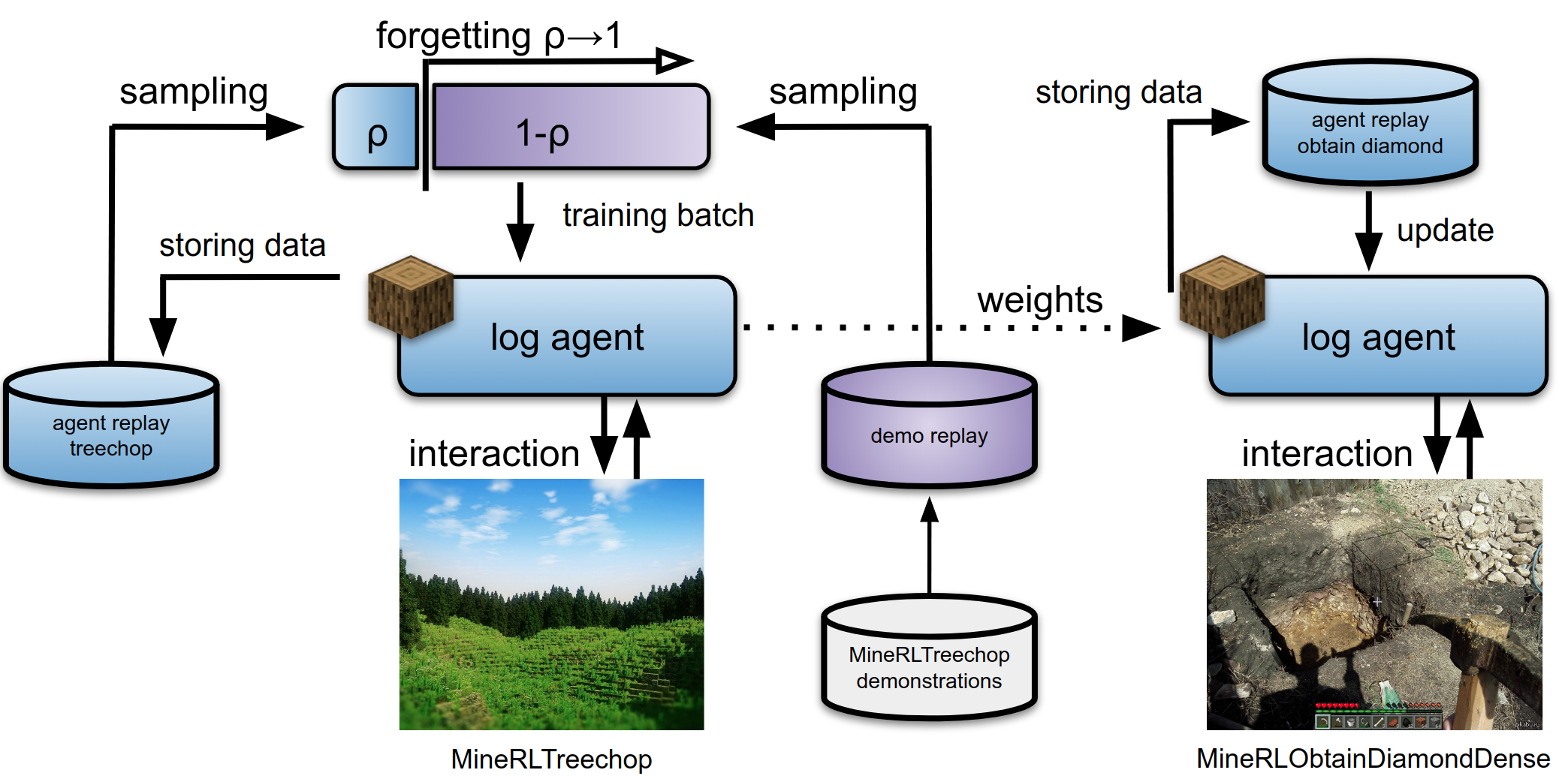}
        \caption{Training \texttt{log} agent.
        The aggregated buffer is used to store expert and agent trajectories.
        The amount of data in the mini-batch sampled from the demo replay buffer is gradually decreasing.}
        \label{fig:agent-schemas-log}
    \end{figure}

    Despite this action, the space discretization allowed making good behavior cloning as there was some noise in the demonstrations due to which the agent could not improve it strategy above a certain threshold.
    We solved this problem by adding an ability to change the ratio of the expert data.
    The demonstrations and the agents' trajectories were stored separately in Aggregating Buffer, which controls the proportion of demonstrations in mini-batches.
    The proportion decreases linearly depending on the number of episodes (see picture~\ref{fig:agent-schemas-log}).

    \subsection*{Extracting hierarchical subtask structure}

\begin{figure}[ht]
        \centering
        \includegraphics[width=0.95\linewidth]{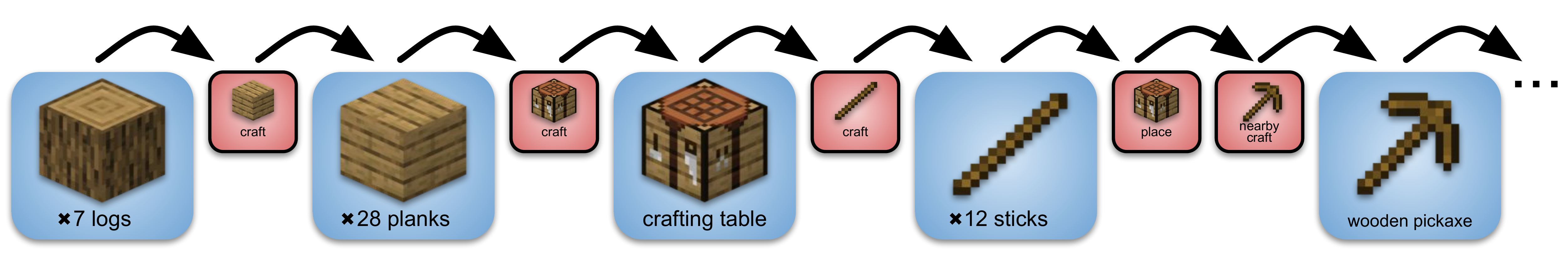}

        \caption{Example of expert's trajectory from demonstrations.
        Action nodes highlighted in red.
        Subtask nodes highlighted in blue.
        Additionally, the number of items for each subtask indicated.}
        \label{fig:chain}
    \end{figure}

    We examined each expert's trajectory separately and considered the time of appearance of the items in the inventory in a chronological order.
    An example of a possible order of obtaining items is shown in~\ref{fig:chain}.
    In addition, this sequence can be considered as a semantic network with two types of nodes: certain agent's actions and subtasks defined for the agent's inventory.
    We consider each subtask node in this network as a mandatory sub-goal that the agent must complete to move on.
    We train a separate strategy for the agent to achieve each sub-goal, and it can be considered as a set of individual agents.
    The task of such agents is to obtain the necessary number of the items in the inventory.

\begin{figure}[ht]
        \centering
        \includegraphics[width=0.9\linewidth]{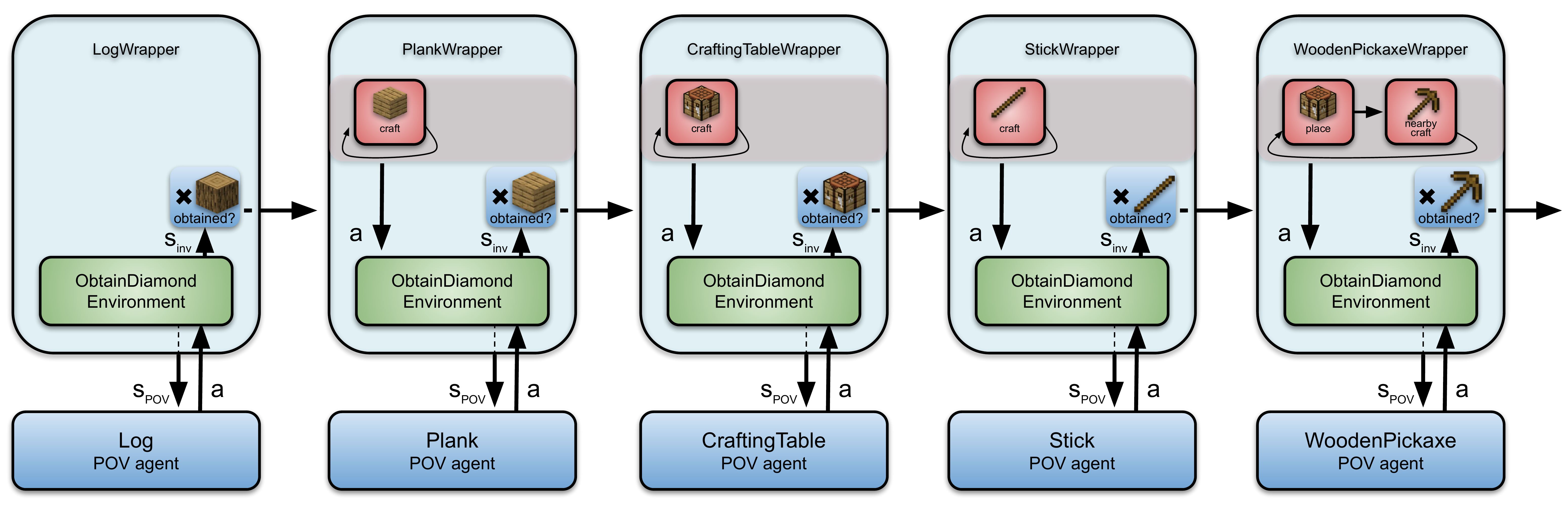}
        \caption{Separation of the agent types (policies) for the subtasks and specific actions to craft and obtain the items.}
        \label{fig:wrapper}
    \end{figure}

\begin{figure}[ht]
        \centering
        \includegraphics[width=0.8\linewidth]{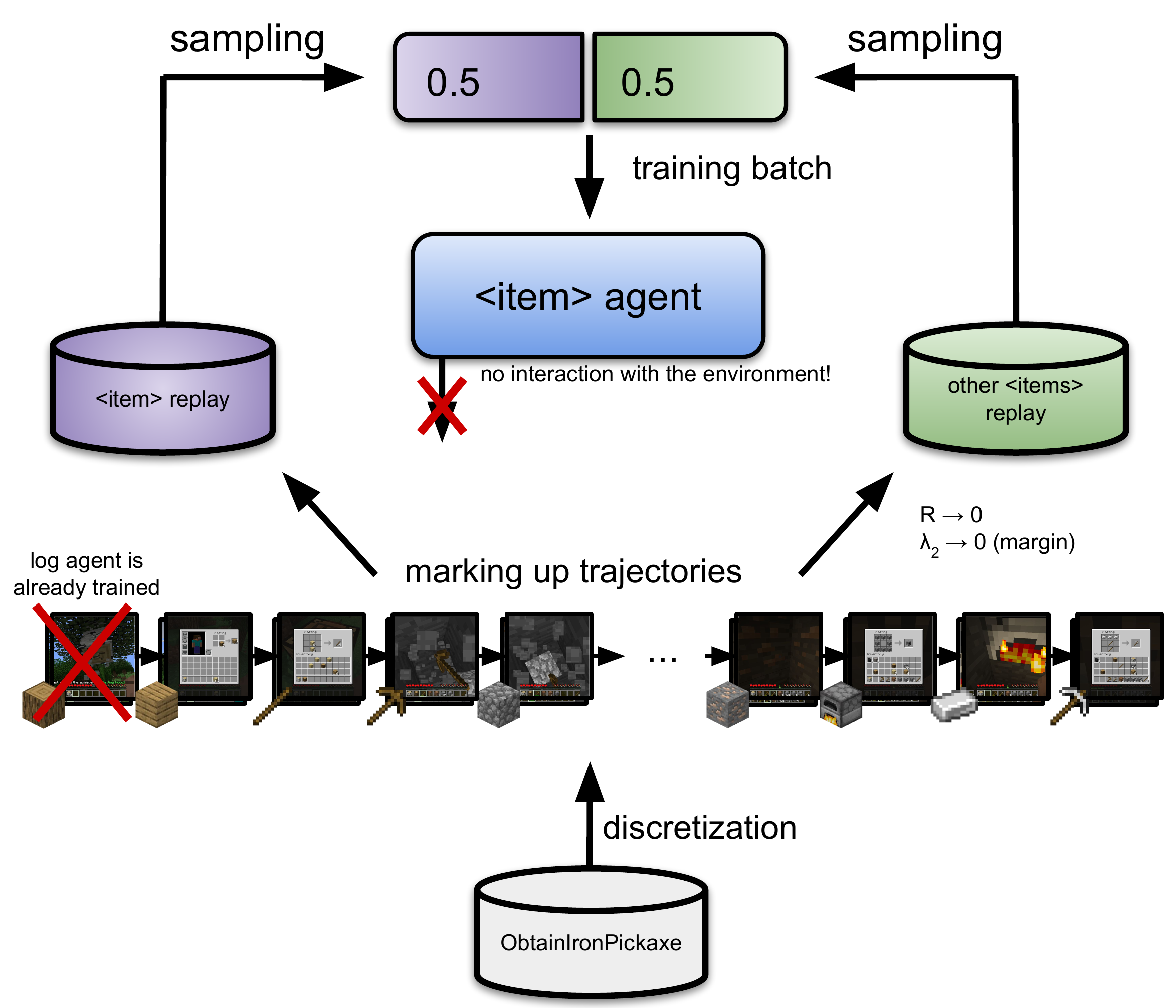}
        \caption{Training \texttt{item} agents.
        For the \texttt{item} agents each trajectory is divided into expert and non-expert segments.
        The item agent learns to solve one subtask using the data from the other subtasks that it considers as non-expert.}
        \label{fig:agent-schemas-item}
    \end{figure}

    The agent that solves the subtask is divided into two agents that take actions at the same time (see~\ref{fig:wrapper}): the agent performing basic actions in the environment (POV or \texttt{item} agent) and the agent interacting with semantic actions -- sequentially perform the action denoted in the corresponding node of the semantic network.
    The training scheme for the \texttt{item} agents is presented in~\ref{fig:agent-schemas-item}.
    During the training process, all the expert data from the \texttt{ObtainIronPickaxe} environment of the MineRL simulator is used.

    The frames of a mini-batch that correspond to the currently trained \texttt{item} agent are considered as expert data.
    All the other frames are considered as additional data and their rewards are nullified.
    This approach allows both training the agent to move from solving one subtask to another in addition to the effective use of available data.

    \section{Experiments}\label{sec:experiments}

    Here we report our results from Round 2 of the MineRL Competition.
    All the agents except for the \texttt{log} agent were trained on the expert data gathered from the \texttt{ObtainIronPickaxeDense} dataset.
    A summary of all the submissions is presented in Table~\ref{tab:round2-submissions}.

    \begin{table}[ht]
        \caption{Round 2 Submissions}
        \label{tab:round2-submissions}
        \centering
        \begin{tabular}{lllllll}
            \toprule
            \multicolumn{5}{c}{Log agent: \texttt{Treechop} $\Rightarrow$ \texttt{ObtainDiamondDense}} \\
            \cmidrule(r){1-5}
            & \texttt{Treechop} episodes & Reward & Episodes & Reward & Pre-training          & Evaluation \\
            \midrule
            Submit 1 & -                          & -      & -        & -      & $10^4$ steps          & 20.72      \\
            Submit 2 & 200                        & 53.22  & 300      & 16.31  & $10^4$ steps          & 55.08      \\
            Submit 3 & 200                        & 53.83  & 300      & 19.19  & $5 \times 10^4$ steps & 61.61      \\
            \bottomrule
        \end{tabular}
    \end{table}

    In the first submit, the HDQfD agent was trained using only expert data.
    Each of the \texttt{item} agents was pre-trained using $10^4$ steps.
    The \texttt{Log} agent learned on the \texttt{Treechop} environment data.
    The final result was \textbf{20.72}.

    In the 2nd and the 3rd submissions we used interaction with the environment to train the \texttt{log} agent.
    The \texttt{log} agent was trained 200 episodes on the \texttt{Treechop} environment data, and then on 300 episodes of the \texttt{ObtainDiamondDense} environment data (see dynamics in the~\ref{fig:submits}).
    The difference was in the number of pre-training steps.
    The final results were \textbf{55.08} and \textbf{61.61}, respectively.
    In addition, in Table~\ref{tab:algos} we provide a comparison of the algorithms that we tested for the \texttt{Treechop} environment.

    \begin{figure}[ht]
        \input{treechop.tex}
        \input{diamond.tex}
        \caption{Log agent results for the \texttt{Treechop} (left) and \texttt{ObtainDiamondDense} environments (right).}
        \label{fig:submits}
    \end{figure}

    \begin{table}[ht]
        \caption{Comparison of the algorithms implemented for the \texttt{Treechop} environment.}
        \centering
        \label{tab:algos}
        \begin{tabular}{llllll}

            \toprule
            & Demonstrations & Discretization & Embeddings & Episodes & Reward \\
            \midrule
            SAC            &                &                &            & 300      & 5      \\
            GAIL           & +              & +              & +          & 150      & 30     \\
            RnD            &                & +              & +          & 1000     & 35     \\
            PPO            &                & +              & +          & 1000     & 35     \\
            Pretrained PPO & +              & +              & +          & 150      & 50     \\
            \textbf{HDQfD} & +              & +              &            & 200      & 60     \\
            \bottomrule
        \end{tabular}

    \end{table}

    \section{Conclusion}\label{sec:conclusion}
    In this paper we introduce a novel approach to learn from imperfect demonstrations.
    This hierarchical Deep Q-Network from Demonstrations won first place in the MineRL competition and received \textbf{61.61} score.
    In our further work, we plan to train all item agents for full hierarchical end-to-end architecture.
    Besides, for these agents we plan to ensure access to all demonstrations from all substasks with respect to the agent's inventory for additional performance.

    \subsubsection*{Acknowledgments}

    This work was supported by the Russian Science Foundation, project no.
    18-71-00143.
    We thank the AIM Tech Company for its organizational and computing support.

    \medskip

    \small
    \bibliographystyle{abbrvnat}
    \bibliography{nips}

\end{document}